\documentclass[letterpaper,twocolumn,10pt]{article}

\usepackage{fullpage}
\usepackage{graphicx,curves}
\usepackage{amsmath}
\usepackage{amssymb}
\usepackage{latexsym}
\usepackage{multirow}

\usepackage[algoruled,vlined,linesnumbered]{algorithm2e}
\usepackage{wrapfig}
\usepackage[hyphens]{url}
\usepackage{comment}
\usepackage[round]{natbib}
\usepackage[symbol]{footmisc}
\usepackage{booktabs}
\usepackage{units}
\usepackage{MnSymbol}
\usepackage{comment}
\usepackage{color}
\usepackage{float}
\usepackage{stfloats}
\usepackage{hyperref}

\newtheorem{df}{Definition}

\newtheorem{theorem}{Theorem}

\newcommand{\bt}{\begin{theorem}\em}
\newcommand{\et}{\end{theorem}}

\newcommand{\bea}{\begin{eqnarray}}
\newcommand{\eea}{\end{eqnarray}}
\newcommand{\bdf}{\begin{df}\em}
\newcommand{\edf}{\end{df}}
\newcommand{\ben}{\begin{enumerate}}
\newcommand{\een}{\end{enumerate}}

\newcommand{\citea}[1]{\citet{#1}}
\renewcommand{\cite}[1]{\citep{#1}}

\renewcommand{\arg}{\operatornamewithlimits{arg}\limits}

\renewcommand{\max}{\operatornamewithlimits{max}\limits}

\numberwithin{equation}{section}
\numberwithin{theorem}{section}
\numberwithin{lemma}{section}
\numberwithin{df}{section}
\begin{document}

\title{Survival Dynamics of Neural and Programmatic Policies in Evolutionary Reinforcement Learning}

\date{}

\author{Anton Roupassov-Ruiz and Yiyang Zuo \\
Department of Computing Science \\ University of Alberta \\
Edmonton, Alberta, T6G 2E8, Canada \\
{\tt roupasso@ualberta.ca} and {\tt zuo5@ualberta.ca}}

\maketitle
\begin{abstract}
In evolutionary reinforcement learning tasks (ERL), agent policies are often encoded as small artificial neural networks (NERL). Such representations lack explicit modular structure, limiting behavioral interpretation. We investigate whether programmatic policies (PERL)---implemented as soft, differentiable decision lists (SDDL)---can match the performance of NERL. To support reproducible evaluation, we provide the first fully specified and \href{https://github.com/antonrou/erl}{open-source re-implementation} of the classic 1992 Artifical Life (ALife) ERL testbed. We conduct a rigorous survival analysis across 4000 independent trials ($\tau_{trial} = 2000$), utilizing Kaplan--Meier curves and Restricted Mean Survival Time (RMST) metrics absent in the original study. We find a statistically significant difference in survival probability between PERL and NERL (log-rank $p \approx 4.1 \times 10^{-72}$). PERL agents survive on average 201.69 steps longer than NERL agents
(95\% CI [180.14, 223.25], $Z = 18.34$, $p \approx 3.8 \times 10^{-75}$).
Moreover, SDDL agents using learning alone (no evolution) survive on average
73.67 steps longer than neural agents using both learning and evaluation (95\% CI [52.8, 94.5], $Z = 6.92$, $p \approx 4.7 \times 10^{-12}$). These results demonstrate that programmatic policies can exceed the survival performance of neural policies in ALife.
\end{abstract}
\footnotetext{Source code available at: \url{https://github.com/antonrou/erl}}

\section{Introduction}

Understanding how learning interacts with evolution has long been a central question in artificial life and adaptive systems \citep{ackley1991interactions, hinton1987how}. 

Although neural policies are compact and expressive, their undifferentiated
weight vectors offer little transparency into the behavioral structures that
evolution discovers. Understanding which behaviors support long-term
survival---and how these behaviors are shaped through learning and
inheritance---remains an open challenge. This gap is further compounded by limitations in the original
Ackley--Littman ERL testbed: despite its lasting influence, the environment
lacks reproducibility (few hyperparameters are reported, learning rates are
unspecified, and no code or formal description of network inputs is provided)
and it lacks quantitative rigor (no survival analysis, statistical testing,
or descriptive statistics are presented for the different neural policy classes compared in the paper). Moreover, the testbed has never been
evaluated using programmatic policies in its partially observable ALife testbed.

Motivated by these gaps, our paper makes the following contributions:

\begin{enumerate}

    \item \textbf{An open-source reimplementation of the Ackley--Littman ERL testbed.}
    We provide the first fully specified and open-source version of the classic
    1991--1992 Artificial Life ERL environment, including all simulation
    parameters, initialization procedures, and evaluation protocols.

    \item \textbf{A programmatic alternative to neural ERL policies.}
    Building on this framework, we investigate whether \emph{programmatic}
    policies can match the survival performance of NERL. Programmatic policies are implemented as
    differentiable soft decision lists: ordered clauses with sigmoid-activated
    conditions whose weighted actions are combined to form a smooth, trainable
    policy. This representation provides an explicit modular structure that is,
    in principle, more amenable to behavioral interpretation than a monolithic
    neural controller.

    \item \textbf{A quantitative survival analysis of neural vs.\ programmatic ERL.}
    We compare NERL and PERL under identical ecological conditions using 4000
    independent trials per policy class. To provide a more rigorous statistical
    analysis than the original testbed, we report Kaplan--Meier survival
    curves, log-rank tests, and RMST analyses.

    \item \textbf{Empirical finding: programmatic ERL outperforms neural ERL in survival.}
    Across all evaluations, we find a statistically significant global difference in survival probability between neural and programmatic policies, suggesting that programmatic rule-based controllers can exceed the long-term
    robustness of neural policies in ALife. We also find that programmatic policies with learning alone can outperform neural policies that combine both learning and evaluation components.

\end{enumerate}


\section{Problem Formulation}
\label{sec:problem}

We model the environment as a bounded \(100 \times 100\) grid world with fixed
edge boundaries, evolving in discrete time steps \(t\). The world contains
five entity types: agents, carnivores, plants (food), trees (shelter), and
walls (obstacles). At each step, carnivores move first, followed by agents.

The task is episodic: each \textit{trial} runs for at most \(\tau_{trial}\) timesteps.
A trial consists of a fixed set of initial conditions that is shared across
\textit{strategies} being compared, and within a trial many generations of
carnivores and agents may be spawned. 

Each strategy specifies a different policy class, and we evaluate strategies using the created policies with survival analysis. Each agent maintains internal state
variables for health \(h \in [0,1]\) and energy \(e \in [0,100]\).

\medskip
\noindent
\textbf{POMDP Formulation.}
From the perspective of each individual agent, the environment forms a 
\emph{Partially Observable Markov Decision Process} (POMDP) \cite{aastrom1965}. This is defined by the tuple
\[
\mathcal{M} = (\mathcal{S}, \mathcal{A}, \mathcal{O}, T, R, O).
\]
We define each component as follows:

\begin{itemize}
    \item \textbf{State space \(\mathcal{S}\).}
    The full environment state includes the positions and types of all entities 
    (Agents, Carnivores, Plants, Trees, Walls).

    \item \textbf{Action space \(\mathcal{A}\).}
    Each agent selects one of four cardinal movement actions 
    \(\mathcal{A} = \{N, E, S, W\}\), represented as two output bits.

    \item \textbf{Observation space \(\mathcal{O}\).}
    Agents receive a local egocentric observation consisting of the nearest 
    entity in each of the four cardinal directions up to the view radius, 
    encoded across six entity types, along with internal proprioceptive 
    features \((h, e, \text{in\_tree}, 1)\). 
    This yields the fixed 28-dimensional observation vector used by both the 
    neural and programmatic strategies.

    \item \textbf{Transition dynamics \(T(s_{t+1} \mid s_t, a_t)\).}
    Transitions are stochastic due to plant growth, carnivore motion, 
    collisions, reproduction events, and the simultaneous actions of all other 
    agents. The global process is Markovian, although transitions are 
    non-stationary from any single agent’s perspective.

    \item \textbf{Reward function \(R(s_t, a_t, s_{t+1})\).}
    Agents receive implicit rewards through survival: movement incurs an energy 
    cost, consuming plants or corpses provides energy, collisions reduce health, 
    and episode termination occurs upon energy or health reaching zero.

    \item \textbf{Observation function \(O(o_t \mid s_t)\).}
    This maps the full environment state to the agent’s partial, egocentric 
    observation. Because only the closest visible entities are reported and the 
    internal states of other agents are hidden, agents operate under significant 
    partial observability.
\end{itemize}

\medskip
\noindent
\textbf{Decision-Making Objective.}
Each agent faces a sequential control problem under partial observability. 
At every time step \(t\), the agent receives an observation 
\(o_t \in \mathcal{O}\), selects an action \(a_t \in \mathcal{A}\) 
according to a policy \(\pi(a_t \mid o_t)\), and the environment evolves 
according to the stochastic transition kernel 
\(T(s_{t+1} \mid s_t, a_t)\). The agent obtains a reward 
\(r_t = R(s_t, a_t, s_{t+1})\) reflecting energy expenditures, food intake, 
and survival. Death occurs when either health or energy reaches zero, 
terminating the individual agent's episode at a random time \(\tau\).

The agent’s objective is to find a policy \(\pi^\ast\) that maximizes 
its expected cumulative survival reward:
\[
\pi^\ast = 
\arg\max_{\pi} \mathbb{E}_{\pi}\!\left[ \sum_{t=0}^{\tau} r_t \right]
\quad = \quad 
\arg\max_{\pi} \mathbb{E}_{\pi}[\tau],
\]
where the second equality holds because the agent receives a constant survival reward \(r_t = r > 0\) at every timestep it remains alive.

\medskip

To evaluate strategies in this ecological POMDP, we formalize survival as the
primary performance signal. Let $\tau_{trial} \in \mathbb{N}$ denote the random
termination time of a trial, i.e., the first time $t$ at which
all agents in a trial have their energy or health reach 0. Survival time therefore serves both as
an implicit reward and as the central performance metric.

\paragraph{Between-Policy Comparison.}
To compare NERL and PERL, we employ
tools from survival analysis:

\begin{itemize}
    \item \textbf{Kaplan–Meier Survival Curve.}
    Let \(T\) be the (random) trial survival time and suppose we observe death times
    \(t_{1} < t_{2} < \dots < t_{J}\).
    For each distinct event time \(t_{j}\), let
    \(d_j\) be the number of deaths at \(t_{j}\) and
    \(n_j\) the number of trials at risk just prior to \(t_{j}\).
    The Kaplan–Meier estimator of the survival function is
    \[
        \hat{S}(t)
        \;=\;
        \prod_{t_{j} \le t}
        \left( 1 - \frac{d_j}{n_j} \right),
    \]
    which yields a step function \(t \mapsto \hat{S}(t)\), commonly plotted
    as the Kaplan–Meier survival curve.
    When comparing two strategies \(A\) and \(B\) with survival curves
    \(\hat S_A(t)\) and \(\hat S_B(t)\), we say that \(A\) has a higher chance of surviving across all timesteps over \(B\) if
    \[
        \hat S_A(t) \;\ge\; \hat S_B(t) \quad \text{for all } t,
    \]
    with strict inequality for some \(t\).
\end{itemize}

\begin{itemize}
    \item \textbf{Log-Rank Test.}
    Given two empirical survival curves
    $\widehat{S}_{A}(t)$ and $\widehat{S}_{B}(t)$, we compute the log-rank
    statistic
    \[
        \chi^{2}
        = \frac{(O_{A} - E_{A})^{2}}{V_{A}},
    \]
    where $O_{A}$ is the observed number of deaths in group $A$, $E_{A}$ is
    the expected number under the null hypothesis of equal hazards, and
    $V_{A}$ is the corresponding variance. Under the null,
    $\chi^{2} \sim \chi^{2}_{(1)}$. This test assesses global differences in
    survival between trials across the full time horizon.
\end{itemize}

\begin{itemize}
    \item \textbf{Restricted Mean Survival Time (RMST).}
    For a fixed time horizon \(\tau_{trial} > 0\), the RMST is defined as
    \[
        \mu(\tau_{trial}) \;=\; \mathbb{E}[\min(T, \tau_{trial})]
        \;=\; \int_{0}^{\tau_{trial}} S(t)\, dt,
    \]
    where \(T\) is the (random) survival time and \(S(t)\) is the survival function.
    In our setting, we seek to \emph{maximize} RMST for each trial and compare strategies based on their RMST values across trials.
\end{itemize}

\paragraph{Evaluation Objective.}
While each agent internally optimizes for maximum expected survival
\[
    \pi^{*} = \arg\max_{\pi} \mathbb{E}_{\pi}[\tau],
\]
our \emph{algorithmic} comparison between policy classes is based on
population-level survival metrics: (i) Kaplan--Meier survival curves $\widehat{S}(t)$,
(ii) the log-rank test for global differences in survival distributions, and
(iii) RMST up to a truncation horizon
$\tau_{trial}$ steps as a time-averaged survival summary.
Taken together, these measures characterize
differences in long-horizon robustness and failure dynamics between the two policy
classes.


\section{Related Work}
\subsection{ERL and the Ackley–Littman Testbed}

The foundational work of \citea{ackley1991interactions} introduced ERL with artificial neural networks (ANNs) in ALife, a hybrid approach in which every agent agent possesses two neural components: an action network that  maps sensory input to behavioral output, and an evaluation network whose weights are genetically inherited and serve as a scalar guiding learning. ERL is embedded in a gridworld featuring carnivores, plants, walls, and trees for shelter. Agents' brains in this setting were either ANNs with both learning and evolution, only learning, only evolution, networks with fixed random weights, or brownian (random walk).

\subsection{Limited Interpretability of Neural Policies in ERL}
\citea{ackley1991interactions} used small, single-layer neural networks for agent policies (for both the action and evaluation networks). While being expressive and allowing ERL populations to signficantly outlive non-ERL populations, a limitation of using neural policies is that they result in limited interpretability of an agent's brain.

\subsection{Lack of Quantitative Evaluation}
A second limitation of \citea{ackley1991interactions} is the lack of a statistically rigorous quantitative evaluation: they do not report variance estimates, confidence intervals, or formal hypothesis tests comparing survival across policy classes.
\subsection{Limited Reproducibility}
Finally, \citea{ackley1991interactions} report limited hyperparameters and architectural decisions, making reproducibility of their results difficult.


\section{Proposed Approach}
A central limitation of the original \citea{ackley1991interactions} testbed is that agent behavior is encoded in neural networks whose weights offer limited transparency into the behavioral structures that evolution discovers. To address this gap, our we propose replacing the neural network with a programmatic policy architecture based on SDDL. This representation consists of ordered clauses that are potentially more interpretable.

\subsection{Input} 

Both PERL and NERL agents observe a $28$-dimensional input feature vector
$\boldsymbol{x} \in \mathbb{R}^{28}$.

\paragraph{Input Vector}
For each direction $d \in \{\text{N},\text{E},\text{W},\text{S}\}$ with index
$k \in \{0,1,2,3\}$ (ordered as N=0, E=1, W=2, S=3), we allocate six scalars corresponding to cell types
\[
\beta \in \{\text{Empty}, \text{Wall}, \text{Plant}, \text{Tree}, \text{Carnivore}, \text{Agent}\},
\]
with type IDs $c \in \{0,\dots,5\}$.
Their indices are
\[
i = 6k + c \quad (0 \le i \le 23).
\]

Along each ray in direction \(d\), we locate the closest cell of type \(c\) at
Manhattan distance \(\texttt{dist}\) (up to \(\texttt{max\_dist = AGENT\_VIEW\_DIST}\)).
We map this distance to a proximity signal via
\[
f(\texttt{dist}) =
\texttt{get\_dist\_value}(\texttt{dist}, \texttt{AGENT\_VIEW\_DIST}),
\]
where \(\texttt{get\_dist\_value}\) is a linear mapping that
\emph{(i)} returns \(0\) if \(\texttt{dist} > \texttt{max\_dist}\),
\emph{(ii)} returns \(1\) at distance \(1\), and
\emph{(iii)} otherwise returns
\[
\texttt{get\_dist\_value}(\texttt{dist}, \texttt{max\_dist})
= 1 - 0.5\,\frac{\texttt{dist} - 1}{\texttt{max\_dist} - 1},
\]
so values are linearly scaled from \(1.0\) (adjacent) to \(0.5\) (at
\(\texttt{max\_dist}\)). The corresponding input feature \(x_i\) is
\[
x_i =
\begin{cases}
f(\texttt{dist}), & \text{if } \texttt{dist} \le \texttt{AGENT\_VIEW\_DIST},\\[2pt]
0, & \text{otherwise.}
\end{cases}
\]

\subsection{NERL}
We reimplement NERL as closely as possible to the original \citea{ackley1991interactions} testbed. The agent's cognitive architecture consists of two single-layer ANNs: an action network that maps sensory input to behavior, and an evaluation network that estimates the goodness of the current state. Both networks observe the input vector $\boldsymbol{x}$. 

\begin{figure*}[!t]
\centering
\includegraphics[width=0.9\textwidth]{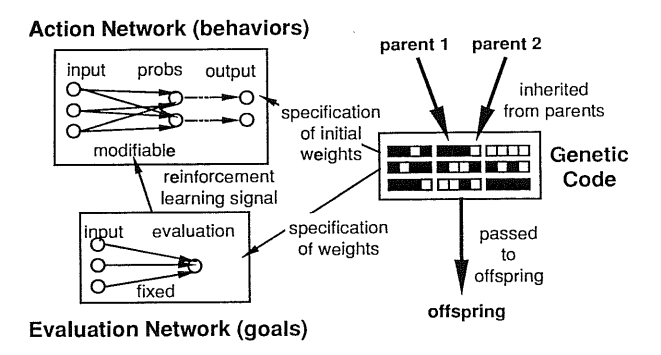}
\caption{Overview of NERL. Reproduced from work by \citea{ackley1991interactions}}
\label{fig:overview-of-erl}
\end{figure*}

Figure~\ref{fig:overview-of-erl} shows that the action network that transforms sensory input into behavioral choices. Its initial weights are inherited, but they adapt over time through reinforcement learning that rewards actions leading to improved evaluations. It has two output neurons.

The evaluation network is genetically determined and converts sensory input into a scalar assessment of situational quality. Its weights never change during the agent's lifetime, so it defines the innate goals that guide learning. It has one output neuron.

\subsection{PERL}
We also reimplement the action network from \citea{ackley1991interactions}
as a soft decision-list policy rather than as a feed-forward single layer neural network.
In principle, one could represent policies as arbitrary programs in a
general-purpose or domain-specific language, but this leads to a
combinatorially large, discrete search space that is difficult to
optimize with gradient-based methods and complicates credit assignment \cite{trivedi2022learningsynthesizeprogramsinterpretable}.
By contrast, a soft decision list provides a rule-structured, fully differentiable parameterization that can be, in principle, more interpretable.

Concretely, each clause consists of a gate, a threshold, and a vector of
action logits. The ordered list of clauses induces a priority structure
analogous to a decision list, and the use of soft, differentiable gates
allows us to train the policy with the same CRBP/REINFORCE-style updates
used in NERL.

\subsubsection{Action Network}
Each clause contains a vector of logits $\mathbf{l}_k$ representing the action preferences if that clause is active. The clause-specific policy is a softmax distribution \cite{Bridle1989}:
\begin{equation}
    \pi_k(a) = \frac{\exp([\mathbf{l}_k]_a)}{\sum_{a'} \exp([\mathbf{l}_k]_{a'})}
\end{equation}
The global policy is the mixture of clause policies weighted by their clause probabilities:
\begin{equation}
    \pi(a\mid \boldsymbol{x}) = \sum_{k=1}^K p(c_k\mid \boldsymbol{x}) \pi_k(a)
\end{equation}

\noindent where \(K\) is the total number of clauses.

\noindent{\paragraph{Gating Mechanism.}
The gate for the $k$-th clause determines the probability that the decision process considers this clause, conditional on it not having been captured by previous clauses. We define the gate activation as:
\begin{equation}
    g_i(\boldsymbol{x})=\sigma(\textbf{w}_{i}^\top\boldsymbol{x}-\tau_i)
\end{equation}
 where $\sigma(x)=\frac{1}{1+e^{-x}}$ and $\boldsymbol{x}=[x_0,x_1,...,x_{27}]^\top$
\smallskip
\citep{Cox1958, Verhulst1977, Kontschieder_2015}.\\}

\paragraph{Clause Probability.}
The probability $p(c_k)$ that clause $k$ is responsible for the final action is calculated by the product of its own gate activation and the probability that all preceding gates $j < k$ failed to trigger (the \textit{else} logic):
\begin{equation}
    p(c_k\mid \boldsymbol{x}) = g_k(\boldsymbol{x}) \prod_{j=1}^{k-1} (1 - g_j(\boldsymbol{x}))
\end{equation}

\subsubsection{Evaluation Network}

To facilitate reinforcement learning within the lifespan of an agent, we implement a secondary structure, the Evaluation Network ($V_\phi$), which mirrors the architecture of the action network. However, instead of action logits, each evaluation clause outputs a scalar utility prediction $u_k$.

The agent's internal estimation of the state value, $e_t$, is computed as:
\begin{equation}
    e_t(\boldsymbol{x}) = \sigma\left(\sum_{k=1}^K p(c_k \mid \boldsymbol{x}) u_k + b_{eval} \right)
\end{equation}
where $b_{eval}$ is a global bias term.

\paragraph{Learning Mechanism: Adapted CRBP.}

We employ a modified version of the Complementary Reinforcement Back-Propagation (CRBP) algorithm in our soft decision list architecture. We also employ a \textit{mental rehearsal loop}, as originally used in NERL \cite{ackley1991interactions}. On positive reinforcement, the reward is applied repeatedly for up to 20 rehearsal iterations, or until another stochastic output generation produces the same result as the initial success; on negative reinforcement, the punishment is applied for up to 20 iterations, or until a new output differs from the initial failure.

The learning signal is derived from the temporal difference in the internal evaluation network:
\begin{equation}
    r = E_t(\boldsymbol{x}_t) - E_{t-1}(\boldsymbol{x}_{t-1})
\end{equation}

We base our learning algorithm on the original REINFORCE algorithm discussed by \citea{williams1992reinforce}, only we take the sign of the reward and not the magnitude. At each timestep \(t\), the environment parameter change \(\Delta\theta\) for the stochastic policy \(\pi_\theta(a\mid s)\) is:
\begin{equation}
\Delta \theta = \alpha \,\operatorname{sign}(r)\,\nabla_\theta \ln \pi(a_t|\boldsymbol{x}_t)
\end{equation}

where $\theta$ is
\[
  \theta = \bigl\{\, (\boldsymbol{w}_k, \tau_k, \mathrm{logits}_k) \,\bigr\}_{k=1}^{K},
\]
\(\alpha\in\mathbb{R}\) is the learning rate factor. Therefore, we are able to update \(\theta\) by using \(\theta\leftarrow\theta+\Delta\theta\).
The parameters $\theta$ (Action Network) are updated to maximize the likelihood of the selected action if $r > 0$, or minimize it if $r < 0$.

\section{Empirical Evaluation}
\begin{table*}[htbp]
\caption{Experimental parameters used in our empirical evaluation.}
\label{tab:experimental_parameters}
\begin{center}
\begin{tabular}{c|c}
\toprule
{\bf Parameter} & {\bf Value} \\
\midrule

\multicolumn{2}{c}{\bf Grid Configuration} \\
\midrule
WORLD\_WIDTH & 100 \\
WORLD\_HEIGHT & 100 \\
\midrule

\multicolumn{2}{c}{\bf Energy \& Health Dynamics} \\
\midrule
MAX\_ENERGY & 100.0 \\
MAX\_HEALTH & 1.0 \\
ENERGY\_COST\_MOVE & 0.3 \\
ENERGY\_GAIN\_PLANT & 100.0 \\
ENERGY\_GAIN\_MEAT & 50.0 \\
DAMAGE\_CARNIVORE & 0.1 \\
DAMAGE\_WALL & 0.1 \\
CORPSE\_DECAY\_RATE & 0.3 \\
\midrule

\multicolumn{2}{c}{\bf Reproduction} \\
\midrule
REPRODUCE\_ENERGY\_THRESHOLD & 60.0 \\
REPRODUCE\_COST & 50.0 \\
\midrule

\multicolumn{2}{c}{\bf Environment \& Spawning} \\
\midrule
PLANT\_GROWTH\_PROB & 0.005 \\
PLANT\_MAX\_DENSITY\_NEIGHBORS & 4 \\
TREE\_BIRTH\_PROB & 0.001 \\
TREE\_DEATH\_PROB & 0.001 \\
CARNIVORE\_SPAWN\_FREQ & 200 \\
\midrule

\multicolumn{2}{c}{\bf Perception} \\
\midrule
AGENT\_VIEW\_DIST & 4 \\
CARNIVORE\_VIEW\_DIST & 6 \\
\midrule

\multicolumn{2}{c}{\bf Neural Agent (ERL)} \\
\midrule
Architecture & Single Layer \\
Learning Rate ($\alpha$) & 0.05 \\
Mutation Rate & 0.05 \\
\midrule

\multicolumn{2}{c}{\bf Programmatic Agent} \\
\midrule
Clauses ($K$) & 2 \\
Learning Rate ($\alpha$) & 0.05 \\
Mutation Rate & 0.05 \\
\bottomrule
\end{tabular}
\end{center}
\end{table*}

\paragraph{Environment.}
Experiments take place in a $100 \times 100$ discrete gridworld populated by Agents, Carnivores, Plants, Trees, and Walls. Boundary walls enclose the world, and additional internal walls are placed randomly (see Table~\ref{tab:experimental_parameters} for full configuration).

\paragraph{Agents.}
Each trial initializes a population of agents with energy, health, and a simple action space. Agents move, forage, and reproduce according to fixed metabolic and reproductive rules. Agents die when either energy or health reaches zero.

\paragraph{Carnivores.}
Carnivores pursue nearby agents within a limited visual range and otherwise move randomly. They regain energy by consuming corpses.

\paragraph{Resources.}
Plants spawn stochastically with density constraints. Trees serve as safe havens: entering a tree makes an agent invisible to carnivores until it climbs down.

\paragraph{Observations and Actions.}
Agents receive a 28-dimensional observation vector combining directional visual sensors and internal state features. Actions are encoded via two output bits corresponding to movement in the four cardinal directions.

\paragraph{Tree Mechanics.}
An agent may enter an unoccupied tree, becoming protected from carnivores but unable to move until climbing down into an adjacent empty cell.

\paragraph{Reproducibility.}
Each experimental batch generates a global seed offset via
\[
\texttt{seed\_offset} = \texttt{int(time.time())} \bmod 10000.
\]
Trial $i$ uses seed $i + \texttt{seed\_offset}$ across all strategies, ensuring identical initial conditions and spawn sequences for fair comparison.

\label{sec:results}

We conduct a quantitative comparison of all policy strategies using survival 
analysis tools. Each strategy is evaluated over 4000 independent trials, with a 
maximum simulation length of 2000 time steps per trial. 

We cap episodes at 2000 timesteps due to hardware constraints; this introduces right-censoring, which is naturally handled by survival analysis and does not invalidate our statistical comparisons.

We generate Kaplan--Meier survival curves \cite{kaplan1958} and use the log-rank (Mantel--Cox) test \cite{mantel1966} and RMST comparisons \cite{royston2013rmst} to assess differences between survival distributions. The strategies evaluated are:
\begin{itemize}
    \item \textbf{NERL}: Neural ERL (full evolutionary reinforcement learning with a neural policy).
    \item \textbf{NE}: Neural Evolution Only (neural policy with evolution but no lifetime learning).
    \item \textbf{NL}: Neural Learning Only (neural policy with lifetime learning but no evolutionary inheritance).
    \item \textbf{NF}: Neural Fixed (neural policy with no learning and no evolution; fixed weights).
    \item \textbf{PERL}: Programmatic ERL (full ERL with a programmatic soft decision-list policy).
    \item \textbf{PE}: Programmatic Evolution Only (evolutionary programmatic policy with no lifetime learning).
    \item \textbf{PL}: Programmatic Learning Only (lifetime learning in a programmatic policy with no evolutionary inheritance).
    \item \textbf{B}: Brownian baseline (agents move by uniform random walk with no learning or evolution).
\end{itemize}

\subsection{Statistical Analyses}

\begin{figure*}[!t]
\centering
\includegraphics[width=0.9\textwidth]{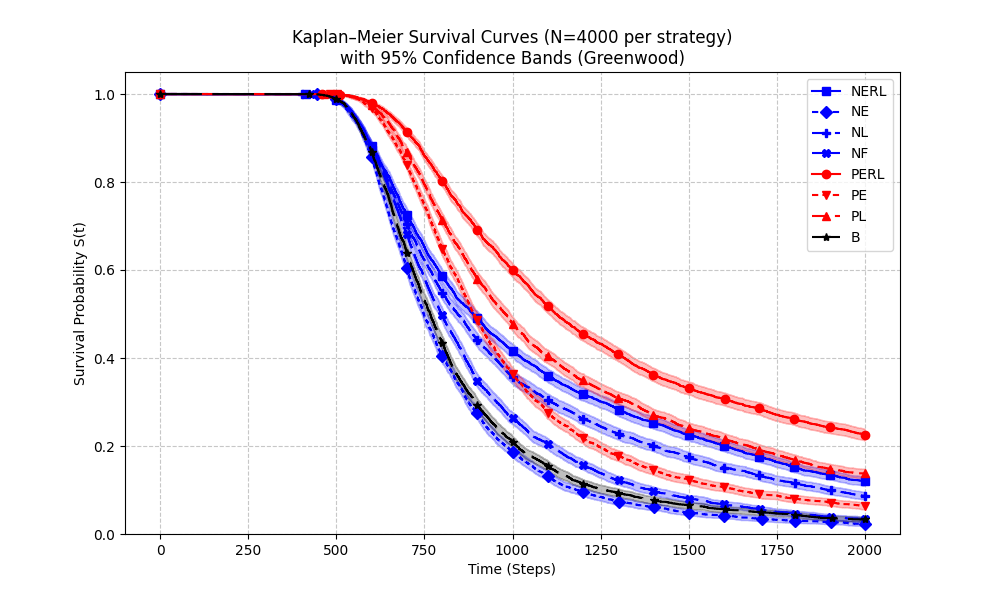}
\caption{Kaplan–Meier survival curves comparing all agent strategies.}
\label{fig:kaplan-meier}
\end{figure*}

\paragraph{Kaplan--Meier Curves.}Figure~\ref{fig:kaplan-meier} shows the Kaplan--Meier survival curves for all eight
agent strategies. PERL has the highest survival probability for any given timestep compared to all other strategies. PL comes in second, followed closely by NERL. PE is third at the start, but later overtaken by both NE and NF. NL, NE, and B have consistently lower survival probabilities compared to the other strategies for any given timestep.

\paragraph{Mean and Median Survival Times.}
Table~\ref{tab:survival-summary} summarizes the average and median survival times, as well as the number of censored trials.

\begin{table}[H]

\centering
\begin{tabular}{lccc}
\hline
\textbf{Strategy} & \textbf{Mean} & \textbf{Median} & \textbf{Censored} \\
\hline
NERL & 1075.5 & 888.0  & 479/4000 \\
NE   & 836.9  & 748.0  & 92/4000  \\
NL   & 1014.6 & 844.0  & 347/4000 \\
NF   & 905.4  & 799.0  & 133/4000 \\
PERL & 1277.1 & 1123.5 & 902/4000 \\
PE   & 1013.0 & 891.0  & 254/4000 \\
PL   & 1149.1 & 980.0  & 551/4000 \\
B    & 863.8  & 766.0  & 132/4000 \\
\hline
\end{tabular}
\caption{Summary of mean, median, and censored data statistics across 4000 trials per strategy ($\tau_{trial} = 2000$).}
\label{tab:survival-summary}
\end{table}

\paragraph{Log-Rank Test.}
To compare NERL and PERL, we compute the
log-rank statistic. For NERL, the observed number of deaths is 3521 and the
expected number of deaths under the null hypothesis is 2800.90. The resulting
test statistic is
\(\chi^2 = 322.5351\).

Using the \(\chi^2\) critical value of 3.841 (df \(= 1\), \(\alpha = 0.05\)), \(p \approx 4.1 \times 10^{-72}\). The log-rank test indicates a statistically significant difference between the survival curves of NERL and PERL at the 5\% significance level, so we reject the null hypothesis of identical survival distributions.

\paragraph{Restricted Mean Survival Time (RMST).}
PERL achieves the highest restricted mean survival time
($\text{RMST} = 1277.14$, SE $= 7.83$), followed by PL
($\text{RMST} = 1149.12$, SE $= 7.34$; see Table~\ref{tab:rmst}). NERL comes in third at ($\text{RMST} = 1075.45$, SE $= 7.72$)
The difference between PERL and NERL is $201.69$ steps
(95\% CI $[180.14, 223.25]$, $Z = 18.34$, $p \approx 3.8 \times 10^{-75}$). This result provides overwhelming evidence that PERL achieves higher restricted mean survival time than NERL at the 5\% significance level.

The difference between PL and NERL is $73.67$ steps
(95\% CI $[52.8, 94.5]$, $Z = 6.92$, $p \approx 4.7 \times 10^{-12}$). This result provides extremely strong evidence that PL achieves higher restricted mean survival time than NERL at the 5\% significance level (see Table~\ref{tab:rmst}).

\begin{table}[t]
\centering
\begin{tabular}{lccc}
\hline
\textbf{Strategy}      & \textbf{RMST}   & \textbf{SE}    & \textbf{95\% CI}              \\
\hline
NERL           & 1075.45 & 7.72 & (1060.3, 1090.6)     \\
NE             & 836.85  & 4.81 & (827.4, 846.3)       \\
NL             & 1014.62 & 7.15 & (1000.6, 1028.6)     \\
NF             & 905.41  & 5.54 & (894.5, 916.3)       \\
PERL           & 1277.14 & 7.83 & (1261.8, 1292.5)     \\
PE             & 1013.02 & 6.01 & (1001.2, 1024.8)     \\
PL             & 1149.12 & 7.34 & (1134.7, 1163.5)     \\
B              & 863.79  & 5.24 & (853.5, 874.1)       \\
\hline
\end{tabular}
\caption{Summary of survival statistics across 4000 trials per strategy ($\tau_{trial} = 2000$).}
\label{tab:rmst}
\end{table}

\section{Discussion}

Using the Kaplan-Meier survival curves, log-rank test results, and RMST, we conclude that PERL achieves better survival performance than NERL. Surprisingly, the RMST results also show that learning alone using our soft decision list architecture can outperform full ERL under a neural policy. This implies that in our ALife testbed, optimizing policy representations can yield larger gains in survival than adding evolutionary machinery to a neural controller, suggesting that architectural inductive bias is a critical driver of performance in ERL.

We additionally observe that, within a neural policy class, learning alone outlives evaluation alone, echoing the findings of \citea{ackley1991interactions}.

Overall, our results show that programmatic policies can surpass neural policies in POMDPs when optimized with ERL. Consequently, even before accounting for their potential interpretability benefits, programmatic ERL with SDDL emerges as a strong alternative to neural ERL.



\section{Future Work}
A natural extension of this work is to increase the simulation horizon beyond the current limit of $2000$ time steps to better probe long-run evolutionary dynamics, albeit at a substantially higher computational cost. Future research could investigate whether the learning--evolution interactions reported at very long horizons (e.g., $> 3,000{, }000$ time steps) in prior ERL work~\cite{ackley1991interactions}---such as the \emph{shielding effect}, where a well-adapted, hard-wired action network effectively insulates a maladapted learning network from selection pressure, and the \emph{Baldwin effect}, where initially learned behaviours (e.g., discovering that plants are beneficial and should be approached) are gradually assimilated into innate action tendencies through ordinary selection---also emerge in PERL.

Future work may also include attempting to interpret an agent’s programmatic policy by studying how its action distribution, realized actions, and the surrounding environment evolve over time.

\section{Conclusions}
We revisited the classic \citea{ackley1991interactions} ERL testbed and addressed several limitations in its original formulation. We provided the fully specified, reproducible, and open-source re-implementation of the environment, enabling statistically grounded comparisons between different strategies. Building on this framework, we investigated whether programmatic policies implemented as differentiable soft decision lists can achieve longer survival than the neural policies traditionally used in ERL.

Empirically, we find that programmatic policies based on soft, differentiable decision lists can outlive neural policies in this complex POMDP, both when the two policy classes are trained with ERL and even in the asymmetric setting where the programmatic policy uses learning alone (no evolution) while the neural policy benefits from both learning and evolution. These findings underscore the role of policy representation in ERL and indicate that structured, programmatic controllers are a competitive alternative to small artificial neural networks, even before accounting for their potential interpretability advantages.


\bibliographystyle{apalike}
\bibliography{project}

\end{document}